# Reconstruction-Based Adaptive Scheduling Using AI Inferences in Safety-Critical Systems

Samer Alshaer, Ala' Khalifeh, and Roman Obermaisser
*Department of Embedded Systems*
*University of Siegen*
Siegen, Germany

*Abstract*—Adaptive scheduling is crucial for ensuring the reliability and safety of time-triggered systems (TTS) in dynamic operational environments. Scheduling frameworks face significant challenges, including message collisions, locked loops from incorrect precedence handling, and the generation of incomplete or invalid schedules, which can compromise system safety and performance. To address these challenges, this paper presents a novel reconstruction framework designed to dynamically validate and assemble schedules. The proposed reconstruction models operate by systematically transforming AI-generated or heuristically derived scheduling priorities into fully executable schedules, ensuring adherence to critical system constraints such as precedence rules and collision-free communication. It incorporates robust safety checks, efficient allocation algorithms, and recovery mechanisms to handle unexpected context events, including hardware failures and mode transitions. Comprehensive experiments were conducted across multiple performance profiles, including makespan minimisation, workload balancing, and energy efficiency, to validate the operational effectiveness of the reconstruction models. Results demonstrate that the proposed framework significantly enhances system adaptability, operational integrity, and runtime performance while maintaining computational efficiency. Overall, this work contributes a practical and scalable solution to the problem of safe schedule generation in safety-critical TTS, enabling reliable and flexible real-time scheduling even under highly dynamic and uncertain operational conditions.

## I. INTRODUCTION

Safety-critical time-triggered systems (TTS) are commonly used in areas like automotive, aviation, industrial automation, and medical devices, where operations must be predictable and reliable. These systems rely on carefully designed schedules that specify exact times for tasks to run and messages to be sent, ensuring deterministic behavior. However, real-world situations can introduce unexpected events such as hardware failures, variations in task execution times (slack), or changes in operational modes. As a result, these systems must adapt quickly and effectively to maintain safety and performance [1] [2].

Metascheduling is a widely adopted solution to provide adaptability in time-triggered systems. Unlike traditional static scheduling, metascheduling involves creating multiple precomputed schedules designed to handle different anticipated scenarios such as hardware failures, task execution slacks, or mode transitions. These schedules are then dynamically selected and deployed in real-time based on the specific context encountered. This approach significantly enhances the flexibility and robustness of TTS, enabling them to maintain reliable operations even under varying and unforeseen conditions. Metascheduling is explored in academic literature under various terminologies, including quasi-static scheduling and super-scheduling [3] [4].

However, metascheduling introduces its own complexities. Although precomputed schedules are highly effective in addressing known events, the sheer number of potential runtime scenarios can be vast, leading to extensive storage requirements. Moreover, precomputed schedules may not always precisely match the actual runtime conditions, necessitating additional dynamic adjustments. This is where reconstruction models become crucial. Reconstruction models bridge the gap between the precomputed priorities from metascheduling and executable schedules, dynamically adjusting and validating schedules to precisely match the real-time conditions [5] [6].

One of the main challenges in adaptive scheduling is turning dynamically determined scheduling priorities into executable schedules that comply fully with all system requirements. Traditional scheduling methods work well when schedules remain fixed, but they often struggle to adapt quickly and reliably to changing conditions. This can lead to issues like message collisions, where simultaneous message transmissions interfere with each other, and locked loops, where tasks become stuck waiting indefinitely due to incorrectly managed precedence constraints. Additionally, schedulers might produce incomplete or invalid schedules that fail to meet the necessary safety and timing constraints required by critical systems. Moreover, ensuring the correctness of dynamically generated schedules is especially crucial because even minor violations of scheduling constraints can have severe implications, including compromised safety or system failures. Therefore, there is a pressing need for mechanisms that can continuously and reliably check the validity of adaptive schedules during runtime, ensuring that any dynamically adjusted schedules remain both safe and executable [7].

To overcome these issues, this paper introduces a new reconstruction framework specifically designed to dynamically convert scheduling priorities into safe and executable schedules. The reconstruction models proposed here check and assemble schedule components, ensuring that messages are sent without collisions and tasks follow the correct order according to their dependencies. This approach directly addresses common scheduling problems such as message interference, locked loops, and incomplete scheduling. While similar reconstruction mechanisms have been referenced in prior literature, they were typically treated as auxiliary components within broader scheduling frameworks and lacked detailed architectural or algorithmic exposition [5] [6] [8] [9].

The core contribution of this research lies in the proposed reconstruction method, which offers a practical solution for validation and adjustment of schedules to ensure safe and reliable system operation. This paper studies the development, analysis, and validation of reconstruction models capable of generating executable schedules focusing on reliability and scalability. Key highlights include guaranteeing collision-free message transmission, robust handling of unforeseen runtime events, and demonstrating scalable applicability across realistic multi-core safety-critical time-triggered system environments.

The remainder of this paper is structured as follows: Section II reviews prior work on adaptive and fault-tolerant scheduling in safety-critical systems. Section III formalises the Application, Platform, and Context Models and motivates the need for schedule reconstruction. The complete framework including our AI inference pipeline and the three reconstruction algorithms is detailed in Section IV. Section V presents an extensive experimental evaluation across makespan, workload-balance, and energy-efficiency profiles, and analyses scalability as well as recovery overhead. Finally, the *Biography* section concludes the paper with brief professional backgrounds of the authors.

## II. LITERATURE REVIEW

The paper published by Kopetz proposes a comprehensive fault hypothesis for the Time-Triggered Architecture (TTA), specifying fault-containment regions, unrestricted failure modes, and recovery intervals to support the design and certification of ultra-dependable distributed embedded systems [10] . Like our reconstruction-based adaptive scheduling scheme, it targets safety-critical time-triggered networks and emphasises explicit fault-tolerance mechanisms that maintain deterministic behaviour under faults. Unlike our work, its assumptions and counter-measures are defined statically at design-time, whereas we employ AI-driven reconstruction models to regenerate feasible schedules online when workload or fault patterns deviate from the original envelope. Accordingly, our approach adds adaptive timing and resource reallocation capabilities to the literature, extending TTA dependability principles to a broader and less predictable operating regime.

Lipari and Palopoli survey how the traditional boundary between hard- and soft-real-time computing is blurring, and unify decades of work on resource-reservation scheduling—especially the Constant Bandwidth Server (CBS)—probabilistic deadline analysis, and adaptive reservations—culminating in a methodology that applies these soft-RT tools to safety-critical control systems without losing certifiability [11]. Like our reconstruction-based adaptive scheduler, their framework tackles timing uncertainty head-on and seeks to preserve predictable behaviour even when workloads or execution times deviate from worst-case assumptions . However, they achieve this by budget-based isolation within an EDF/event-triggered OS (e.g., Linux SCHED˙DEADLINE), leaving the original static task set intact, whereas we regenerate an entire time-triggered schedule online through AI-driven reconstruction to keep deterministic bus arbitration and message ordering intact under faults or mode changes. Consequently, our work extends the literature by bringing soft-RT adaptability into certified Time-Triggered Architecture via machine-learning-guided schedule reconstruction, broadening fault coverage while retaining the temporal determinism required for ultra-dependable networks.

Ornée et al. present a context-aware status-updating framework for safety-critical wireless sensor networks that quantifies a "situation-unaware penalty" combining the Age of Information with the latest sensor value, then minimises this penalty via a Restless Multi-Armed Bandit formulation and an asymptotically-optimal "Net-Gain Maximisation" scheduler [12]. Like our AI-driven reconstruction scheduler, their work tackles uncertain operating conditions in safety-critical domains by adaptively deciding when and how to transmit or schedule tasks to keep system knowledge fresh and dependable . However, they operate in a pull-based, event-triggered wireless setting where tasks remain static and only channel access is re-allocated, whereas we regenerate an entire deterministic time-triggered schedule on-line—preserving global bus arbitration and precedence when faults or mode changes occur. Consequently, our contribution extends their situational-awareness philosophy into certified Time-Triggered Architectures by coupling the same "awareness-driven" responsiveness with full schedule reconstruction, thereby unifying soft-RT adaptability with hard-RT determinism.

Edinger's dissertation proposes Tasklet, a lightweight, context-aware computation-offloading framework that wraps code and data into self-contained "tasklets," schedules them via a brokered peer-to-peer overlay, and optionally enforces Quality-of-Computation (QoC) guarantees such as redundancy or heartbeat-based monitoring to survive volatile edge resources . Similar to our reconstruction-based adaptive scheduler, Tasklet observes run-time context (provider stability, reliability, load) and selects resources or replans to uphold application-level goals under failures [13]. The key difference is that Tasklet operates in a best-effort, event-triggered cloud/edge environment where determinism is optional, whereas our work rebuilds an entire time-triggered schedule runtime—preserving global bus arbitration and certification constraints demanded by ultra-dependable distributed embedded systems. Consequently, our contribution extends Edinger's context-aware philosophy into certified Time-Triggered Architectures by adding an AI-driven reconstruction layer that delivers hard-real-time determinism alongside adaptive fault coverage.

Li et al. introduce a network-wide run-time reconfiguration framework for Time-Triggered (TT) Ethernet/TSN networks, coupling four complementary strategies (local, global, elastic and degraded) with a joint-ILP formulation (JILP) that simultaneously remaps applications, reroutes flows and regenerates schedules after core or link failures [14]. Like our AI-driven reconstruction scheduler, their approach aims to keep safety-critical TT traffic deterministic while coping with unexpected resource faults, and both exploit global knowledge (via an SDN controller in their case) to decide new mappings and schedules. The difference is that they rely on deterministic optimisation (JILP) plus a heuristic (SCA) grounded in "scheduling compatibility" metrics rather than machine-learning models, achieving about a 50× speed-up over plain ILP yet still executing offline to pre-compute options. Our work extends this line by using learning-based reconstruction models to generate feasible TT schedules online, pushing adaptation further into real-time and handling workload drifts that lie outside any pre-enumerated ILP solution set.

Behera presents a fault-tolerant time-triggered scheduling algorithm for uniprocessor mixed-criticality systems, layering "backup" slices onto the baseline TT-Merge tables so the schedule can survive at most one transient fault within each hyper-period [15]. Like our reconstruction-based adaptive scheduler, Behera's work explicitly targets safety-critical TT environments and guarantees that hard-real-time tasks still meet their deadlines even when a run-time fault or Worst Case Execution Time (WCET) overrun occurs. However, Behera's scheme is purely analytic and offline—pre-computing multiple static tables and assuming a single-core platform—whereas we use ML-guided online reconstruction to regenerate a fresh global schedule across distributed TTA/TSN networks whenever arbitrary faults or workload drifts arise. Consequently, our work extends the state of the art by eliminating the "single-fault, single-core, pre-computed" constraint and providing fast, AI-driven re-scheduling that preserves deterministic bus arbitration and certification arguments in large-scale systems.

Saravanaguru and Thangavelu propose CoMiTe, a context-aware middleware that couples OWL-Time ontology with a Context-Aware Time-Petri-Net (CATPN) model to predict situations, schedule actions, and guarantee timely driver-safety responses in distributed vehicular networks (InVANET) [16]. Like our AI-driven reconstruction scheduler, CoMiTe continually senses run-time context in safety-critical environments and adapts in real time to maintain deadline-sensitive behaviour. However, its adaptations stay at the middleware/event-notification layer—leaving the underlying task set and network timetable fixed—whereas we regenerate the entire deterministic time-triggered schedule online with learning-based models to preserve certification-grade temporal guarantees. Thus, our work extends CoMiTe's predictive, context-aware philosophy into the realm of certified Time-Triggered Architectures, fusing temporal semantics with machine-learning-guided schedule reconstruction for broader fault and workload coverage.

Table I presents a comparison of existing adaptive scheduling approaches across nine important features: runtime reconstruction, ML compatibility, network reconfiguration, mixed-criticality support, scalability, certification alignment, real-time overhead handling, and context awareness. While earlier systems like Kopetz's time-triggered architecture and Behera's mixed-criticality scheduler provide deterministic operation and fault tolerance, they do not support runtime flexibility or newer techniques. Lipari Palopoli proposed resource-based scheduling methods but these are mostly offline and do not include machine learning or dynamic features. The approaches by Ornée et al. and Edinger focus on context awareness and adaptability, but they lack scalability, full network reconfiguration, machine learning integration, and alignment with certification needs. Li et al. support reconfiguration and mixed-criticality using optimization techniques, but still do not cover learning, certification, or real-time execution overhead. Saravanaguru Thangavelu's CoMiTe adds context-aware middleware in vehicle systems but does not include learning-based or certified scheduling. In contrast, our proposed framework supports all nine features, combining real-time operation, learning-based decision-making, and full system adaptability for use in safety-critical environments.

## III. BACKGROUND

This section elaborates on the foundational concepts of metascheduling and highlights the essential role of reconstruction models in translating dynamically adjusted scheduling priorities into executable and constraint-compliant schedules. Initially, it provides detailed explanations of the core system components—including the Application Model (AM), Platform Model (PM), and Context Model (CM)—that interact closely with reconstruction mechanisms. Understanding these components and their interactions is crucial for comprehending how the proposed reconstruction framework ensures adaptability and operational integrity in dynamic, safety-critical environments.

### A. Application Model

The AM in our system serves as a comprehensive representation of the tasks that need to be scheduled. It is a critical component of the metascheduling framework, providing all necessary details about each task, including its dependencies, execution requirements, and communication needs. In our implementation, the AM is represented as a series of dataframes, which are structured to contain specific information about the tasks and messages within the system.

The Task Dataframe is central to the Application Model, encapsulating various attributes of each task that are crucial for the scheduling process. The key fields within the Task Dataframe include:

- Task ID: a unique identifier for each task within the system.
- Parents: a list of task IDs that must be completed before this task can begin, representing task dependencies.
- Children: a list of task IDs that are dependent on the completion of this task, indicating tasks that follow in sequence.

TABLE I: Comparison of Literature Works Using Domain-Specific Performance Metrics

| Work | Runtime Schedule Reconstruction | ML/AI- Generated Priority Compatibility | Network- Wide Reconfiguration | Mixed- Criticality Support | High- Scale Scalability | Certification Alignment | Real-Time Computation Overhead | Context Awareness |
|---|---|---|---|---|---|---|---|---|
| [10] | | | | | X | X | X | |
| [11] | X | | | X | X | X | | |
| [12] | X | | | | X | | | X |
| [13] | X | | | | X | | | X |
| [14] | | | X | X | X | X | | |
| [15] | | | | | | | | |
| [16] | | | | | X | | | X |
| **Ours** | X | X | X | X | X | X | X | X |

- WCET: the maximum time required to complete the task under worst-case conditions, which is critical for ensuring that deadlines are met.
- Message size: The size of the data that needs to be transmitted between tasks, which impacts the communication load and influences scheduling decisions.

Table III is an example table representing the structure of the task dataframe:

TABLE III: Task Dataframe Example

| Task ID | Parents | Children | WCET (ms) | Message Size (KB) |
|---|---|---|---|---|
| 1 | - | 2, 3 | 10 | 2 |
| 2 | 1 | 4 | 15 | 4 |
| 3 | 1 | 5 | 20 | 3 |
| 4 | 2 | - | 25 | 6 |
| 5 | 3 | - | 30 | 5 |

From Table III it is noticed that for Task 1, there are no parent tasks, indicating it is the starting task. Task 2 and Task 3 are dependent on Task 1. Task 2 leads to Task 4, and Task 3 leads to Task 5, highlighting the flow of execution. Task 1 has a WCET of 10ms, which means it must be completed within this timeframe under worst-case conditions and the amount of data (in KB) that Task 1 needs to transmit is 2 KB.

In addition to the Task Dataframe, the Application Model also includes a Message Dataframe. This dataframe specifically handles the communication between tasks, detailing the data transfer requirements that influence the scheduling process. The key fields within the Message Dataframe include:

- Transmitting Task ID: The ID of the task that is sending the message.
- Receiving Task ID: The ID of the task that is receiving the message.
- Message Size: The size of the message being transmitted. This structured representation allows the system to efficiently manage task dependencies and communication. Table IV provides an overview of the key elements in the message dataframe, focusing on the relationship between transmitting and receiving tasks along with the size of the messages exchanged between them. You can use this table to complement the explanation of how tasks communicate within your system [17].

TABLE IV: Message Dataframe Example

| Transmitting Task ID | Receiving Task ID | Message Size (KB) |
|---|---|---|
| 1 | 2 | 2 |
| 1 | 3 | 3 |
| 2 | 4 | 4 |
| 3 | 5 | 5 |
| 4 | 6 | 6 |
| 5 | 7 | 8 |

*B. Platform Model*

In this paper, the PM is represented through data structures that capture the details of the hardware resources, including routers, links, and end systems. These resources are crucial for executing tasks efficiently and ensuring robust communication across the network. The PM is implemented using dataframes and lists, which are structured to store and organise information about each end system and its connections to other end systems. The End System dataframe contains key information about each end system, such as its identifier, capabilities, and the associated routes that define its connections with other end systems. Routes are represented as paths consisting of a sequence of routers and links, facilitating communication between end systems. This organised representation allows the scheduling algorithms to make informed decisions regarding task allocation and data transmission, optimising the overall system performance [17]. An example of a hybrid network topologies used in this paper as PM is illustrated in Figure 1. The system is specifically designed to effectively distribute tasks among end systems. The end systems have a distributed memory architecture. Additionally, it facilitates message passing through three routers R1, R2 and R3. The strategic selection and management of network topologies play a critical role in network design, as they significantly impact the overall effectiveness and adaptability of communication.

Table V provides an example of the data captured in the PM through dataframes, specifically focusing on the routes between different end systems. Each row in the table represents a connection between two end systems, including the route that data must traverse, defined by the sequence of routers and links. In table V:

- Sender End System ID: this column lists the ID of the

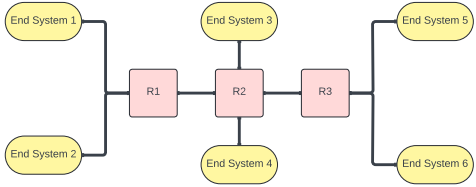

Figure 1: Connection between processors and routers.

TABLE V: Example Routes Between End Systems

| Sender End System ID | Receiver End System ID | Route |
|---|---|---|
| 1 | 3 | R1 → R2 |
| 1 | 4 | R1 → R2 |
| 1 | 5 | R1 → R2 → R3 |
| 2 | 3 | R1 → R2 |
| 2 | 6 | R1 → R2 → R3 |
| 3 | 5 | R2 → R3 |
| 4 | 6 | R2 → R3 |
| 5 | 6 | R3 |

end system that is initiating the communication.
- Receiver End System ID: this column identifies the end system that is intended to receive the communication.
- Route: the route specifies the path that the message or data packet will take through the network. The path is defined by the sequence of routers (e.g., R1, R2, R3) that the data must traverse to reach its destination.

### C. Context Model

The CM is a crucial component of the system, encapsulating all relevant information regarding context events that influence the scheduling process. While these events have been briefly introduced in previous sections, this section will delve deeper into their significance, mechanics, and impact on the overall system performance. The Context Model not only monitors these events but also guides the dynamic adaptation of schedules in response to changes in the operational environment. This section will provide a detailed exploration of various context events, including their triggers, types, and how they interact with other models within the metascheduling framework [18].

*1) Slacks:* in scheduling slack refers to the amount of time that a task can be delayed without affecting the overall completion time of the schedule, often referred to as the makespan. Slacks provide flexibility within the schedule, allowing for adjustments in the order or timing of tasks. Properly managing slacks is crucial in real-time systems, as it ensures that tasks can be reallocated or rescheduled without violating their deadlines allowing for less schedule makespan. Slacks are particularly useful in scenarios where certain tasks complete earlier than their WCET, creating an opportunity to advance other tasks or reallocate resources more efficiently.

Figure 2 illustrates slack handling in a scheduling system across three end systems (ES1, ES2, and ES3). (a) shows the initial schedule where task T0 on ES1 finishes earlier

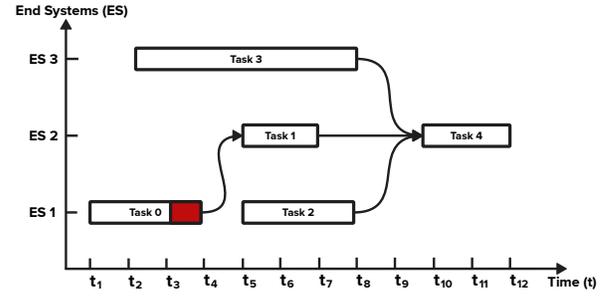

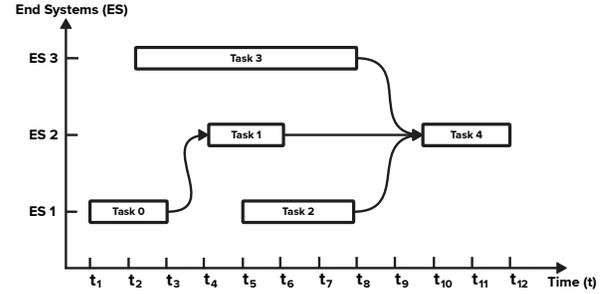

Figure 2: Slack Management in metascheduling. (a) shows the schedule before the slack event occurrence, (b) shows the schedule after the slack event occurrence

than its WCET, creating a slack period, marked in red. (b) demonstrates the adjustment of the schedule to handle slacks, which shifts subsequent tasks to utilise the slack, leading to an earlier makespan and a more efficient schedule.

*2) Hardware Failure:* In the context of scheduling and resource management, a hardware failure refers to an unexpected malfunction or breakdown of any physical component within the system, such as processors, memory units, network interfaces, or communication links. These failures can disrupt the normal operation of the system, leading to delays, missed deadlines, or complete task failures.

When a hardware failure occurs, the system must promptly detect the issue and invoke a contingency plan to maintain operational integrity. This typically involves the Metascheduler reassigning tasks that were originally allocated to the failed hardware components to other available resources. The goal is to ensure that the system can continue functioning with minimal disruption, albeit with potentially reduced performance or efficiency. The Metascheduler uses real-time monitoring data and predictive models to assess the impact of the failure and make decisions on task reallocation. Additionally, the system might trigger reconfiguration processes, such as rerouting communication paths or activating redundant hardware components, to mitigate the effects of the failure.

Effective handling of hardware failures is critical in safety-critical systems, where maintaining continuous operation is essential. The ability to adapt to such failures without significant performance degradation or system downtime is a key feature of robust scheduling frameworks.

*3) Mode Change:* A mode change in a scheduling system refers to the transition between different operational states or modes of the system. These changes can be triggered by various factors, such as shifts in operational requirements, environmental conditions, or the need to conserve energy. For example, a system might transition from a high-performance mode to a low-power mode to reduce energy consumption or switch to a fault-tolerant mode in response to detected hardware failures. In this paper, mode changes have been specifically studied in the context of three key modes: energy efficiency mode, workload balance mode, and performance mode. Each mode has unique operational objectives, such as minimizing energy consumption, balancing the workload across resources, or optimising overall system performance. During a mode change, the system's tasks, resource allocations, and scheduling priorities may need to be adjusted to align with the new mode's requirements. This transition is managed by the Metascheduler, which dynamically reconfigures the schedule based on the current mode. It re-evaluates task priorities, reallocates resources, and modifies execution timelines to ensure the system continues to meet its deadlines and performance goals. Mode changes can significantly impact how tasks and resources are allocated, as each mode prioritises different aspects of system performance. For instance, switching to an energy efficiency mode might reduce the frequency of task execution to conserve power, requiring tasks to be rescheduled to maintain critical operations. In contrast, a workload balance mode might redistribute tasks more evenly across available resources to prevent bottlenecks. These mode changes and their effects on scheduling and resource allocation will be explored in greater detail later in this paper. Understanding these dynamics is crucial for designing adaptable and efficient scheduling systems that can operate under varying conditions.

*D. Scheduling System Overview*

The scheduling unit is tasked with processing inputs from the AM, PM, and CM. These inputs are then provided to the AI scheduling inferences, which use the extracted information to generate temporal and spatial priorities. The AI inferences play a crucial role in determining key parameters to generate schedules by analysing the input data and producing priorities that guide the execution order and resource allocation for tasks within the system.

As depicted in Figure 3, the operation of the scheduling unit is initiated by the online operation manager, which is responsible for aggregating crucial data from the AM, PM, and CM through the system information gathering block. This block serves as the interface between the different models and the AI scheduling inferences, ensuring that the most relevant information is extracted for processing.

The collected data from the AM, PM, and CM encompass various parameters that reflect the current state of applications, platform resources, and contextual conditions. This information forms the input to either the temporal AI inferences or the spatial AI inferences, both of which are trained offline using the Multi Schedule Graph (MSG). The training process equips these inferences with the capability to analyse input data and generate temporal or spatial priorities, tailored to optimise the scheduling process.

Once the AI inferences have processed the input data, they generate temporal and spatial priorities, which are essential for determining the order of task execution and the allocation of resources. These priorities are then passed on to the reconstruction model, which synthesises them into schedules. The reconstruction model integrates these priorities to form a holistic schedule that ensures effective task execution, taking into account the dynamic nature of the system's operational environment.

## IV. IMPLEMENTATION

This section presents a detailed implementation that underpins the proposed system architecture, explaining how different components interact within the framework. It details the structure of the system, emphasising the relationship between application, platform, and context models with a metascheduler that supports a variety of scheduling algorithms, including GNN, ANN, E/D NN, and RFC. The section also examines the function of online learning units, leveraging RL algorithms and neural network predictors to enhance adaptability in real-time scenarios. Additionally, it discusses reconstruction models that aid in the dynamic generation and assessment of schedules, ensuring the system's effectiveness in complex operational contexts. This overview establishes a foundation for a more detailed examination of each component's role in achieving task scheduling in challenging environments.

*A. Context Handling*

The fault management process within a TTS is depicted in Figure 4, where the x-axis denotes the time steps ($t$) and the y-axis illustrates the end systems executing tasks (T1, T2, T3,..., T8). The figure encapsulates two principal phases in context event management.

An intermediate scheduling phase is activated to mitigate the impact of the context event, illustrated by a system failure at End System 3 at t=9. This phase extends from t=9 to t=14, during which tasks continue to be executed in an adjusted schedule that compensates for the disrupted end system. Subsequent to this phase, at t=14 in Figure 4, a revised schedule is implemented. This new scheduling framework accommodates the absence of End System 3 by reallocating tasks to alternative end systems, ensuring continuity and system resilience. Ordinarily, this process is executed only when a failure occurs.

The mathematical representation of these interactions can be illustrated through the following equations, encapsulating the modifications induced by context events:

$$AM' = AM \oplus \Delta AM(CM_{\text{slack}}) \tag{1}$$

Here, $AM'$ denotes the modified application model, $\oplus$ symbolises the update operation, $\Delta AM$ represents the changes dictated by the context model specific to slacks, and $CM_{\text{slack}}$ indicates the slack information within the CM.

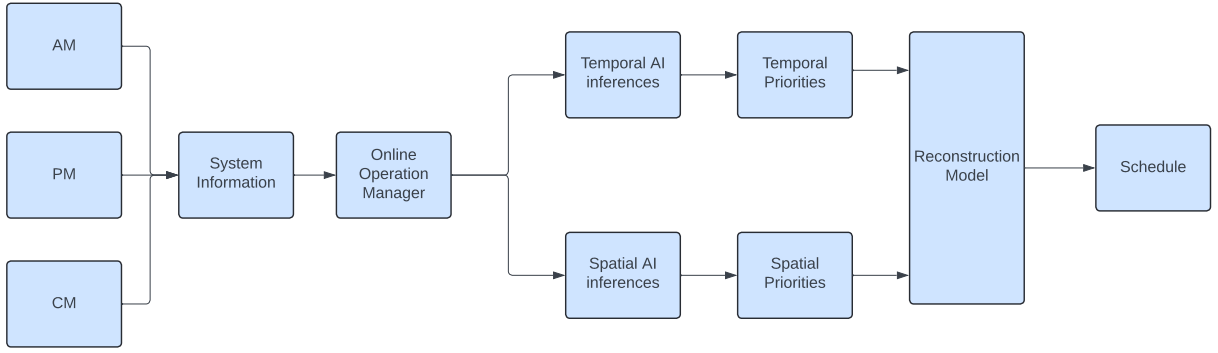

Figure 3: Component of the scheduling unit [19]

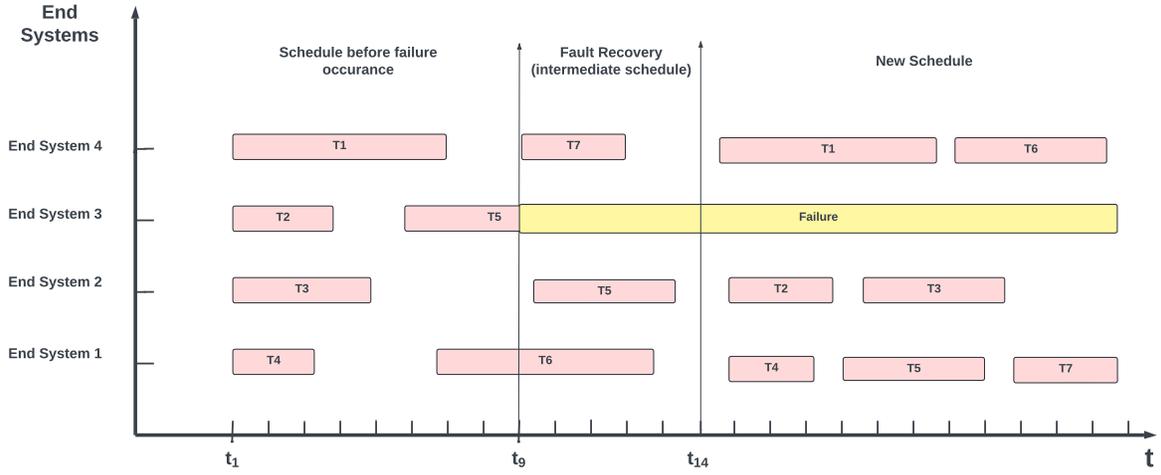

Figure 4: Fault handling in time-triggered systems [19]

$$PM' = PM \oplus \Delta PM(CM_{\text{failure}}) \qquad (2)$$

In this equation, $PM'$ is the updated platform model, $\oplus$ stands for the modification operation, $\Delta PM$ encapsulates the adjustments required by the platform model due to failures, and $CM_{\text{failure}}$ refers to the failure details provided by the CM.

$$\begin{aligned} AM', PM' = {} & (AM \oplus \Delta AM(CM_{\text{mode}})), \\ & (PM \oplus \Delta PM(CM_{\text{mode}})) \end{aligned} \qquad (3)$$

This equation represents the scenario where a mode change affects both the AM and PM, necessitating comprehensive updates to ensure system integrity and functionality. Here, $CM_{\text{mode}}$ highlights the modifications within the CM that influence both models.

These equations conceptualise how contextual changes are seamlessly integrated into the system's operational models, ensuring robust adaptability and sustained system performance. The equations formulated for the system are designed to manage multiple context events occurring simultaneously.

### B. Reconstruction Models

The reconstruction model plays a pivotal role within the overarching system architecture, serving as a critical component that translates the decisions made by the scheduling inferences into actionable schedules. The reconstructor takes the outputs from the online operation manager, which include temporal priorities, spatial priorities, and context event timings, to assemble a coherent and executable schedule. It ensures that all operational constraints, such as message collisions and precedence requirements, are met while also integrating safety checks.

The design and functionality of different reconstruction models vary; some are tailored for comprehensive schedule rebuilding, while others focus on rapid response to immediate contextual changes. This section delves into the various aspects of reconstruction models, detailing their integration with the system's dynamic environment and their critical role in maintaining system resilience and adaptability.

The primary components of the reconstruction model are depicted in Figure 5. The generation of scheduling information, which includes temporal and spatial priorities, necessitates the integration of the AM, PM, and CM. Priorities may be derived from basic built-in algorithms or more sophisticated sources, such as ML models or GA. The built-in algorithm uses bottom level to determine the temporal priorities of a task, taking into consideration WCET and their precedence constraints, as shown in Equation 4:

$$b\text{-level}(v) = w(v) + \max_{(v \to u) \in E}(b\text{-level}(u)) \quad (4)$$

Where:
- $b\text{-level}(v)$ represents the bottom-level value of task $v$, which is an estimate of the longest path from task $v$ to any exit task in the graph.
- $w(v)$ denotes the execution time or the weight of task $v$.
- $\max_{(v \to u) \in E}(b\text{-level}(u))$ calculates the maximum b-level among all tasks $u$ that are direct successors of $v$ in the task graph, where $E$ represents the set of edges indicating dependencies from task $v$ to task $u$.
- Task Weight ($w(v)$): This is the computation time required to complete the task $v$.
- Precedence Constrains: The maximum b-level among all successors ensures that the longest path from the task $v$ through any of its successor paths is considered [20].

As for spatial priorities, when the built-in algorithm is employed, it uses a search algorithm for the least loaded end system to deploy tasks.

In the event of a context that necessitates fault recovery, relevant details such as the event timing, affected end systems, affected tasks, and modifications to the WCET are extracted from the CM.

Recovery variables are stored from the reconstruction model, which log the internal variables of the reconstructor over time. These variables are loaded at the moment hardware failure occurs, fixing tasks executed prior to the context event and adjusting information for tasks occurring after the context event saving significant time caused by the re-computation of new variables in the case hardware failure occurs.

For the generation of new schedules $AM'$ and $PM'$ are extracted, which are derived by integrating updates into the original AM, PM, and CM. These modifications reflect changes induced by context events, transforming the original AM and PM accordingly. These transformations are detailed in Equations 4.1, 4.2, and 4.3, demonstrating how dynamic context elements systematically influence the scheduling framework.

The schedule reconstructor performs several critical functions:
- **Task Allocation**: Assigns tasks to appropriate end systems based on priority information and system capacities.
- **Message Allocation**: Manages data transfers between tasks while preventing message collisions.
- **Fixing Past**: In fault recovery scenarios, fixes the parameters of tasks executed prior to the context event and reallocate only those tasks that are pending, as detailed in previous sections.
- **Safety check**: The reconstructor ensures precedence constrains and handles message collisions.
- **Schedule Generation**: Ultimately, it constructs an update schedule.

Finally, an evaluation block is employed to generate performance metrics tailored to assess the schedules based on various profiles and evaluation criteria. When changing profiles, essentially, this is the only block that changes in the reconstructor. The evaluation block would calculate either makespan, energy efficiency or workload of the produced schedule.

In this paper, reconstructors are termed based on fault recovery and selector settings. Specifically, there are three distinct types of reconstruction models: failure recovery reconstructors, temporal recovery reconstructors, and new schedule reconstructors.

*1) Temporal Recovery Reconstruction:* Algorithm 1 shows the temporal recovery reconstructor. The temporal recovery reconstructor recovers temporal priorities from the AI scheduling inferences and executes a minimal algorithm for spatial priorities allocating tasks to the available end system with the least load. This reconstruction model is designed to test the operation of AI temporal priority schedulers and generate schedules that use these temporal priorities.

The parameters of the algorithm are as follows:
- $task\_data\_list$: This parameter holds the complete list of tasks in the system. Each task is detailed with its specific requirements, dependencies, and originally scheduled times.
- $Message\_data\_list$: Contains all the communication requirements between tasks, including the expected messages, their sizes, and the sender and receiver information.
- $end\_systems$: An array or list detailing all the end systems (processors) available for task allocation. This helps in determining where tasks can be scheduled.
- $context\_event\_time$: This is the timestamp or the point in the system's operational timeline when the context event occurs. Tasks scheduled before this time are not altered, whereas those scheduled after are subject to rescheduling based on the new system state.

The algorithm begins by marking all tasks that were completed before the context event. These tasks are "locked", and their scheduling is preserved to ensure system consistency. Tasks and messages that are scheduled after the context event are then identified for re-evaluation. Tasks are analysed for their readiness based on unresolved dependencies that might have been affected by the context event. This step is crucial for understanding which tasks can be immediately rescheduled and which must wait for other tasks to complete. Each task is allocated to an end system. This allocation is based on which processor is currently least utilised, thereby aiming to balance the load across all end systems. The goal is to minimise the waiting time for any task and maximise overall utilisation. For each task, the communication cost (time delay due to transferring data between tasks) is calculated. The algorithm also checks for potential collisions on communication paths between tasks. Collision management ensures that no two tasks interfere with each other's data transfer, preserving the integrity and reliability of task execution. As tasks are scheduled, conflicts may arise due to concurrent access to resources. The algorithm dynamically adjusts task start and end times to resolve these conflicts, using management of processor time and task dependencies. After all tasks have been reassigned after the context event, the algorithm recalculates the makespan for the new schedule. Makespan is the total time required to complete all scheduled tasks, and minimising it is a primary goal of scheduling. The final step involves the execution of the Allocatetasks procedure in algorithm 1, which applies all the aforementioned scheduling logic to the tasks and messages affected by the context event. By efficiently handling task rescheduling in the face of unpredicted events, algorithm 1 ensures that the system can remain operational and adapt to changes without significant downtimes. The algorithm's focus on conflict resolution and communication management is particularly vital in environments where system integrity and data accuracy are paramount.

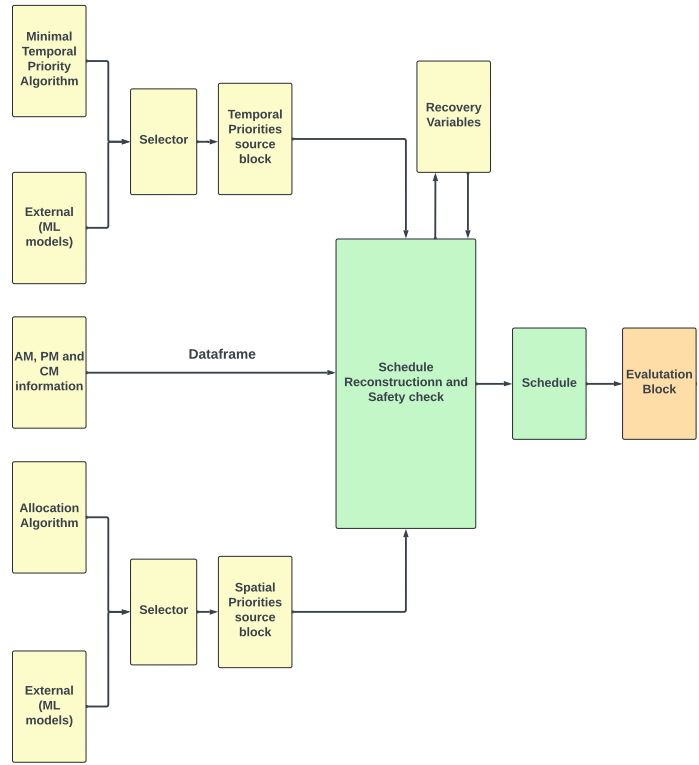

Figure 5: Building blocks of reconstructors [19]

Temporal recovery reconstruction proves useful while handing slack events. As slack events rarely requires spatial reallocation of tasks, temporal reconstruction that affects the timing execution of tasks is used.

*2) Fault Recovery Reconstruction:* Algorithm 2 shows the failure recovery reconstructor or the minimal reconstructor. The algorithm is designed to handle the dynamic scheduling of tasks in a system after a hardware failure. This algorithm ensures that any tasks processed before the context event retains its scheduling decisions, thus focusing only on those tasks that are affected by or scheduled after the event. The failure recovery reconstructor is designed only to preform intermediate stage scheduling. That would mean in the case of a hardware failure the current schedule is inoperable, because it would try to execute tasks on end systems that are inaccessible . At the time of failure occurrence, a one-time swiftly calculated intermediate schedule is fired to manage the completion of the tasks that have not been yet executed as shown in Figure 4.

The failure recovery reconstructor relies on previously stored internal recovery variables from previous reconstruction operations, that would imply that it does not start calculating schedules from scratch. Additionally, it uses minimal allocation techniques for spatial and temporal priorities. The parameters of the algorithm are as follows:

## Algorithm 1 Temporal Recovery Reconstruction Algorithm

**Require:** *task_data_list*, *message_data_list*, *end_systems*, *context_event_time*
**Ensure:** *finalTable*, *makespan*

/* Initialisation */

1: Initialise processor end times, task readiness, communication paths, and collision lists for each router
2: **procedure** ALLOCATETASKS(*task_data_list*, *message_data_list*, *end_systems*, *context_event_time*)
3:    Identify tasks executed before *context_event_time* and lock their schedule
4:    Extract tasks and messages scheduled after *context_event_time*
5:    Initialise task readiness based on dependency resolution
6:    Initialise lists to manage task dependencies and processor allocations
7:    **for all** $t$ in tasks scheduled after *context_event_time* **do**
8:       Determine the processor with minimal load and free resources
9:       Calculate communication costs and paths for task dependencies
10:      Adjust task start/end times and check link–collision lists
11:      Update end-system times and task readiness
12:      Resolve dependencies and update scheduling order
13:      Monitor and resolve concurrent-resource conflicts
14:      Update router-collision lists and manage message passing
15:    **end for**
16:    Assess and calculate the new *makespan*
17:    Update *finalTable* to reflect post-event adjustments
18: **end procedure**
19: ALLOCATETASKS(*task_data_list*, *message_data_list*, *end_systems*, *context_event_time*)

## Algorithm 2 Failure Recovery Reconstruction Algorithm

**Require:** *task_data_list*, *message_data_list*, *recovery_variables*, *context_event_time*, *context_event_type*
**Ensure:** *finalTable*, *makespan*

1: Load *recovery_variables* to retrieve state up to *context_event_time*
2: Identify and lock tasks executed before *context_event_time* as immutable
3: Adjust system state according to *context_event_type* (e.g. failure, slack, mode change)
4: **procedure** GENERATEINTERMEDIATESCHEDULE(*tasks*, *messages*, *context_event_time*, *context_event_type*)
5:    Determine impact of *context_event_type* on end-systems and WCETs
6:    Reconfigure task dependencies and readiness for the new context
7:    Initialise data structures for post-context allocation and messaging
8:    **for all** $t$ in tasks scheduled after *context_event_time* **do**
9:       Calculate new scheduling priorities
10:      Optimise processor allocation to minimise makespan and balance load
11:      Manage communication paths to avoid collisions and ensure integrity
12:      Update task start/end times to honour new constraints
13:      Recalculate communication delays; adjust schedule accordingly
14:      Re-assess and resolve resource conflicts arising from context change
15:    **end for**
16:    Compute and store the new *makespan*
17:    Update *finalTable* with the revised schedule
18: **end procedure**
19: GENERATEINTERMEDIATESCHEDULE(*task_data_list*, *message_data_list*, *context_event_time*, *context_event_type*)

- $task\_data\_list$: Contains detailed listings of all tasks including their dependencies, resources required, and initial scheduling.
- $Message\_data\_list$: Details the communication requirements between tasks, including message sizes and intended paths, crucial for managing data flow between tasks.
- $Recovery\_variables$: Stores comprehensive snapshots of the system state at every time unit, providing a robust foundation for recovery and rescheduling.
- $Context\_event\_time$: The exact time at which the context event occurs, used to demarcate which tasks are locked and which are subject to rescheduling.
- $Context\_event\_type$: Defines the nature of the context event, whether it be a failure, a slack adjustment, or a mode change, each affecting the system in uniquely impactful ways.

The procedure *Generate Schedule* takes these inputs and uses them to:

1) Adjust the operational parameters of the system based on the type of context event.
2) Re-evaluate and reschedule tasks that occur after the context event time, integrating new system constraints and requirements.
3) Ensure that the system maintains integrity and efficiency despite the disruptions, aiming to minimise the overall makespan while adhering to the new operational guidelines.

**Algorithm 3** New Schedule Generation Reconstruction Algorithm

**Require:** Modified Application Model $AM'$ (Task ID, WCET, Parents, Children, Message Size); Platform Model $PM'$ (End Systems, Routers, Connections); Messages DataFrame; Spatial allocation priorities; Temporal allocation priorities

**Ensure:** Schedule DataFrame & updated recovery variables for each iteration

1: Initialise the Schedule DataFrame and recovery variables
2: **procedure** ALLOCATETASKS($AM'$, $PM'$, Messages, SpatialPriorities, TemporalPriorities)
3:    **for all** task $t$ in $AM'$ **do**
4:       Determine optimal end-system for $t$ using $PM'$ and *SpatialPriorities*
5:       Schedule $t$ according to *TemporalPriorities*, preserving all parent/child constraints
6:       Calculate message travel time and detect potential collisions
7:       **if** $t$ involves cross-end-system communication **then**
8:          Adjust timing of $t$ for message latency and router availability
9:       **end if**
10:      Update the Schedule DataFrame with start/end of $t$
11:      Log current system state into *recovery_variables*
12:    **end for**
13:    Check and resolve any remaining precedence or safety violations
14: **end procedure**
15: **return** the final Schedule DataFrame

This algorithm not only ensures high system resilience but also adapts to changes with minimal impact on overall system performance, which is crucial for maintaining service levels in dynamic and demanding operational environments.

*3) Entire Schedule Reconstruction:* Algorithm 3, plays a critical role in the system architecture by dynamically allocating tasks based on predetermined priorities and system constraints. The primary inputs to this algorithm include the modified application model $AM'$, which details task-related data such as execution time and communication needs, and the modified platform model $PM'$, which describes the computational resources and network topology. Additionally, spatial and temporal priorities, which may be derived from internal algorithms or external AI inferences, guide the task allocation process. New schedule reconstruction would not require fixing the settings of past events as does not require context event timing as input. The reason behind that is that it generates complete schedules from scratch.

The procedure begins by initialising a schedule DataFrame, which will ultimately record the start and end times for each task, ensuring that all tasks are executed within their respective time frames without violating system constraints such as message collisions or precedence requirements. Each task is then individually processed to determine the most suitable end system for execution based on spatial priorities. Spatial priorities are provided by the built-in allocation algorithm or from an external source (AI inference). Temporal priorities are used to refine the scheduling order, aligning task execution with optimised system performance criteria.

During the task allocation phase, the algorithm also considers inter-task communications, adjusting task timings to accommodate message travel times and router availability to prevent data collisions. This ensures system integrity and operational safety.

Moreover, the algorithm continuously updates recovery variables throughout the scheduling process. These variables capture the system's state at each iteration, providing a robust mechanism for fault recovery by allowing the system to revert to a previously known good state in the event of a failure.

This proactive approach to scheduling and system recovery underscores the algorithm's utility in maintaining high system performance and reliability, even in complex operational environments.

## V. RESULTS AND CONCLUSION

This section introduced a wide variety of reconstruction algorithms that perform safety checks and schedule reconstruction from temporal and spatial priorities across different profiles. A series of tests were conducted to ensure that reconstruction algorithms do not exceed timing requirements and perform in an optimised manner.

Figure 6 presents a comparison of average runtimes for the spatial reconstruction model across different performance profiles: makespan, average workload, and energy consumption. The x-axis represents the number of tasks, while the y-axis, plotted on a logarithmic scale, represents the average runtime in seconds. This graph helps to illustrate how the complexity of the reconstruction model affects runtime as the number of tasks increases.

The blue line with circle markers depicts the average runtime for the makespan profile. The runtime increases steadily from approximately $1.19 \times 10^{-6}$ seconds for 5 tasks to $5.06 \times 10^{-4}$ seconds for 50 tasks. The makespan profile, which aims to minimize the total time required to complete all tasks, shows a relatively moderate increase in runtime, indicating its efficiency in handling a growing number of tasks.

The red dashed line with square markers represents the average runtime for the average workload profile. This profile demonstrates a more pronounced increase in runtime, starting from approximately $3.80 \times 10^{-5}$ seconds for 5 tasks and escalating to $1.88 \times 10^{-2}$ seconds for 50 tasks. The steeper rise in runtime suggests that balancing the workload across tasks imposes a greater computational burden on the reconstruction model.

The green dash-dotted line with triangle markers indicates the average runtime for the energy consumption profile. The energy profile exhibits the most significant increase in runtime, beginning at approximately $2.10 \times 10^{-5}$ seconds for 5 tasks and soaring to $2.66 \times 10^{-2}$ seconds for 50 tasks. This substantial rise underscores the complexity of optimizing energy consumption, which likely involves more intricate calculations and adjustments to achieve efficient energy usage.

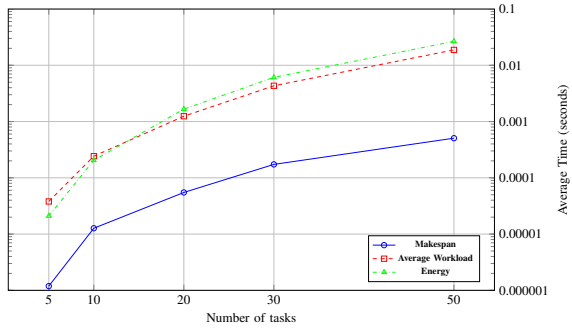

Figure 6: Comparison of average runtimes for spatial reconstruction model in different profiles

Overall, the graph reveals that while all profiles experience increased runtimes with a higher number of tasks, the energy consumption profile incurs the highest computational cost, followed by the average workload profile, and then the makespan profile. These insights highlight the varying computational demands of different optimization objectives within the spatial reconstruction model, emphasizing the need for tailored strategies to manage runtime efficiency based on the specific performance criteria.

Figure 7 shows the minimal b_level built-in algorithm for producing different temporal priorities. The b_level algorithm consumes more time the more complicated the scheduling problem becomes.

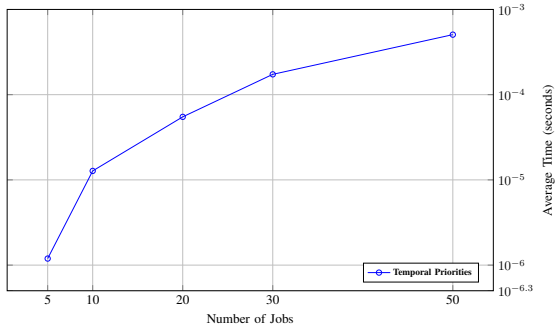

Figure 7: b_level performance with time

The graph in Figure 8 illustrates the relationship between the number of tasks and the required space for recovery variables in the failure recovery reconstructor. As the number of tasks increases, there is a noticeable increase in the required space for recovery variables.

For a small number of tasks, such as 5, the required space is relatively minimal at 2.58 KB. However, as the number of tasks increases to 15, there is a significant jump to 60.0 KB. This trend continues with 30 tasks requiring 112 KB and 50 tasks needing 174 KB. The most substantial increase is observed with 100 tasks, which require 918 KB of space.

These observations suggest that the space requirement for recovery variables grows non-linearly with the increase in the number of tasks. This exponential growth indicates that as the system scales up, more substantial storage resources will be necessary to accommodate the recovery variables, which could impact the overall system performance and resource allocation. Careful consideration and planning are essential to ensure that the system can handle the increased demands efficiently as the number of tasks grows.

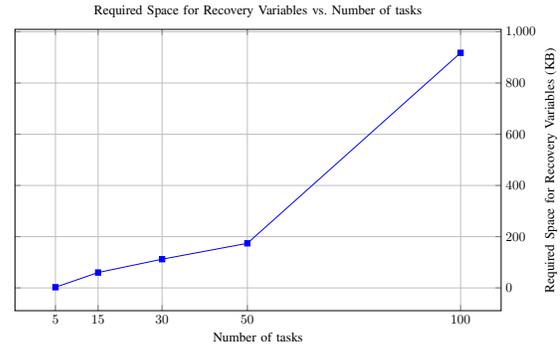

Figure 8: Required Space for Recovery Variables of the Failure Recovery Reconstructor for Different Numbers of tasks

Figure 9 illustrates the average recovery time for the failure recovery reconstructor as a function of the number of tasks. The data points represent the average recovery time sampled over 1000 runs for each task size.

From the graph, it is evident that there is a significant increase in average recovery time as the number of tasks increases. For a small number of tasks, such as 5, the average recovery time is relatively minimal at approximately 0.00018 seconds. However, as the number of tasks increases to 15, the average recovery time more than triples to approximately 0.00071 seconds. This trend continues with 30 tasks requiring an average recovery time of approximately 0.00336 seconds, and 50 tasks needing around 0.01557 seconds.

The rapid increase in recovery time for higher numbers of tasks suggests that the failure recovery reconstructor's efficiency diminishes as the system load increases.

These observations highlight the need for optimising the failure recovery reconstructor to handle larger task sizes more efficiently. As the number of tasks grows, the system's ability to recover from failures in a timely manner becomes increasingly critical. Future improvements could focus on parallel processing techniques or more efficient algorithms to mitigate the exponential growth in recovery time and ensure robust performance under higher system loads.

Figure 10 illustrates the average runtime for the paralleled reconstructor as a function of the number of tasks. The x-axis represents the number of tasks, while the y-axis shows the average runtime in seconds. This graph demonstrates how the implementation of parallel computing techniques affects the execution time of the reconstructor.

13 processing threads were used to conduct this experience. The results indicate a consistent and linear increase in runtime as the number of tasks grows. For 5 tasks, the average runtime is approximately 0.0021 seconds. As the number of tasks increases to 10, the runtime nearly doubles to about 0.0041 seconds. This trend continues linearly with 20 tasks requiring an average runtime of approximately 0.0131 seconds, 30 tasks taking around 0.0210 seconds, and 50 tasks resulting in an average runtime of about 0.0389 seconds.

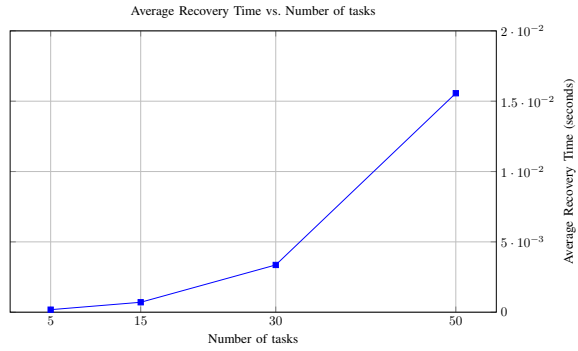

Figure 9: Average recovery time for failure recovery reconstructor

The graph suggests a linear increase in runtime, indicating that the parallelisation has effectively managed to scale the execution time in proportion to the number of tasks. This linear relationship suggests that the paralleled reconstructor is efficiently distributing the workload across multiple processors, maintaining a consistent performance increase as the task count rises.

In conclusion, the use of parallel computing techniques has successfully resulted in a linear increase in execution time for the reconstructor, making it highly scalable and efficient for larger numbers of tasks. This linear scalability is a significant improvement over a potentially exponential increase that could occur without parallelisation. Continued use and optimisation of these parallel techniques will ensure robust performance even as system demands grow.

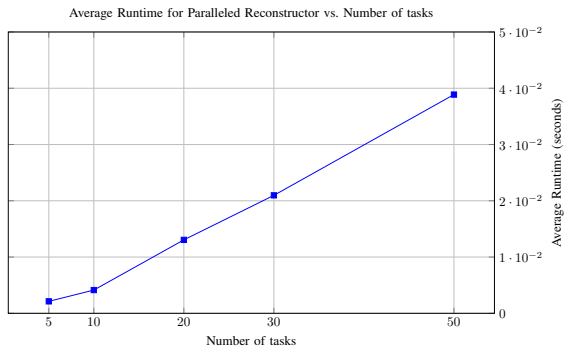

Figure 10: Average runtime for paralleled reconstructor with different numbers of tasks

## VI. Acknowledgement

One of the authors, S. Alshaer, gratefully acknowledges the support of the German Academic Exchange Service (DAAD) for funding his PhD studies, which enabled his contribution to this work. The authors also thank the University of Siegen's Omni Cluster for providing the computational resources used in the data generation process.

## References

[1] B. Sorkhpour, "Scenario-based meta-scheduling for energy-efficient, robust and adaptive time-triggered multi-core architectures," 2019.

[2] J. Cortadella, A. Kondratyev, L. Lavagno, A. Taubin, and Y. Watanabe, "Quasi-static scheduling for concurrent architectures," *Fundamenta Informaticae*, vol. 62, no. 2, pp. 171–196, 2004.

[3] N. Kadri and M. Koudil, "A survey on fault-tolerant application mapping techniques for network-on-chip," *Journal of Systems Architecture*, vol. 92, pp. 39–52, 2019.

[4] M. L. M. Peixoto, M. J. Santana, J. C. Estrella, T. C. Tavares, B. T. Kuehne, and R. H. C. Santana, "A metascheduler architecture to provide qos on the cloud computing," in *2010 17th International Conference on Telecommunications*, 2010, pp. 650–657.

[5] S. Alshaer, C. Lua, P. Muoka, D. Onwuchekwa, and R. Obermaisser, "Graph neural networks based meta-scheduling in adaptive time-triggered systems," in *2022 IEEE 27th International Conference on Emerging Technologies and Factory Automation (ETFA)*, 2022, pp. 1–6.

[6] S. Alshaer, M. Piridi, and R. Obermaisser, "Model comparative analysis of neighborhood aggregation levels in graph neural networks for metaschedulers," in *2024 IEEE International Conference on Industrial Technology (ICIT)*, 2024, pp. 1–7.

[7] A. Yeolekar, R. Metta, C. Hobbs, and S. Chakraborty, "Checking scheduling-induced violations of control safety properties," in *International Symposium on Automated Technology for Verification and Analysis*. Springer, 2022, pp. 100–116.

[8] D. Onwuchekwa, M. Dasandhi, S. Alshaer, and R. Obermaisser, "Evaluation of ai-based meta-scheduling approaches for adaptive time-triggered system," in *2023 International Conference on Smart Computing and Application (ICSCA)*, 2023, pp. 1–8.

[9] C. Lua, D. Onwuchekwa, and R. Obermaisser, "Ai-based scheduling for adaptive time-triggered networks," in *2022 11th Mediterranean Conference on Embedded Computing (MECO)*. IEEE, 2022, pp. 1–7.

[10] H. Kopetz, "The fault hypothesis for the time-triggered architecture," in *Building the Information Society: IFIP 18th World Computer Congress Topical Sessions 22–27 August 2004 Toulouse, France*. Springer, 2004, pp. 221–233.

[11] G. Lipari and L. Palopoli, "Real-time scheduling: from hard to soft real-time systems," *arXiv preprint arXiv:1512.01978*, 2015.

[12] T. Z. Ornee, M. K. C. Shisher, C. Kam, and Y. Sun, "Context-aware status updating: Wireless scheduling for maximizing situational awareness in safety-critical systems," in *MILCOM 2023-2023 IEEE Military Communications Conference (MILCOM)*. IEEE, 2023, pp. 194–200.

[13] J. Edinger, *Context-aware task scheduling in distributed computing systems*. Universitaet Mannheim (Germany), 2019.

[14] J. Li, H. Xiong, Q. Li, F. Xiong, and J. Feng, "Run-time reconfiguration strategy and implementation of time-triggered networks," *Electronics*, vol. 11, no. 9, p. 1477, 2022.

[15] L. Behera, "A fault-tolerant time-triggered scheduling algorithm of mixed-criticality systems," *Computing*, pp. 1–23, 2022.

[16] R. K. Saravanaguru and A. Thangavelu, "Comite: Context aware middleware architecture for time-dependent systems: A case study on vehicular safety," *Arabian Journal for Science and Engineering*, vol. 39, pp. 2895–2908, 2014.

[17] P. Muoka, D. Onwuchekwa, and R. Obermaisser, "Adaptive scheduling for time-triggered network-on-chip-based multi-core architecture using genetic algorithm," *Electronics*, vol. 11, no. 1, p. 49, 2021.

[18] R. Obermaisser, H. Ahmadian, A. Maleki, Y. Bebawy, A. Lenz, and B. Sorkhpour, "Adaptive time-triggered multi-core architecture," *Designs*, vol. 3, no. 1, p. 7, 2019.

[19] S. Alshaer, A. Khalifeh, and R. Obermaisser, "Adaptive approach to enhance machine learning scheduling algorithms during runtime using reinforcement learning in metascheduling applications," 2025, manuscript submitted for publication; under review at *IEEE Transactions on Parallel and Distributed Systems*.

[20] M. Bhatti, C. Belleudy, and M. Auguin, "Two-level hierarchical scheduling algorithm for real-time multiprocessor systems," *JSW*, vol. 6, pp. 2308–2320, 11 2011.

## VII. Biography Section

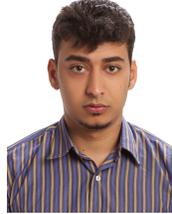

**Samer Alshaer** is currently a PhD candidate in Computer Engineering at the University of Siegen, Germany (since 2021), where he also works as a Research Assistant in the Embedded Systems division. He is a recipient of the DAAD research scholarship and winner of the QRCE Industry-Academia Linkage Competition. Prior to his doctoral studies, Samer served as a Research Assistant at the German Jordanian University (GJU). He earned his M.Sc. in Computer Engineering from GJU in 2019 and graduated first of his class with a B.Sc. in Mechatronics Engineering from Philadelphia University in 2014. His research focuses on intelligent scheduling systems, machine learning, reinforcement learning, and time-triggered architectures. He has authored multiple peer-reviewed publications in these fields.

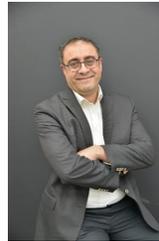

**Dr. Ala' Khalifeh** currently holds the position of President's Advisor for University Campus Life and diversity at the German Jordanian University (GJU), where he also works as a Professor of Electrical Engineering . Dr. Khalifeh received the prestigious Fulbright Scholarship in 2005, which enabled him to pursue his doctoral degree from the University of California-Irvine in the United States of America. In September 2021, Dr. Khalifeh became a Fellow of the Innovation Leaders Fellowship (LIF) Program run by the Royal Academy of Engineering - UK. Dr. Khalifeh's research focuses on emerging technologies including the Internet of Things, artificial intelligence, datal analytics, cloud computing and wireless sensor networks. Recognizing Dr. Khalifeh's research impact, he recently was recognized by Stanford-Elsevier Ranking of Top 2% Top-cited Researchers Includes in 2024.

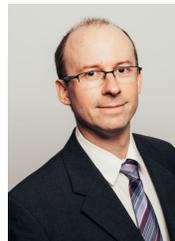

**Prof. Dr.-Ing. Roman Obermaisser** received the master's and Ph.D. degrees in computer sciences from the Vienna University of Technology, in 2001 and 2004, respectively, under the supervision of Prof. Hermann Kopetz. He was a Research Advisor with the Vienna University of Technology. He is currently a Full Professor with the Division for Embedded Systems, University of Siegen. He wrote a book on an integrated time-triggered architecture published by Springer-Verlag, USA. He is the author of several journal articles and conference publications. He has also participated in numerous EU research projects, such as SAFEPOWER, universAAL, DECOS, and NextTTA. He was the Coordinator of European research projects, such as DREAMS, GENESYS, and ACROSS. In 2009, he received the Habilitation ("Venia Docendi") Certificate for technical computer science.